\documentclass[10pt, a4paper]{article}

\usepackage{url}
\usepackage{lrec2016}
\usepackage{multibib}
\usepackage{amsmath}
\usepackage{multirow}
\usepackage{color}
\DeclareMathOperator*{\argmax}{\arg\!\max}
\usepackage{graphicx}

\usepackage{verbatim}
\usepackage{epstopdf}
\usepackage[latin1]{inputenc}

\title{A Large-Scale Multilingual Disambiguation of Glosses}  

\name{Jos\'e Camacho-Collados, Claudio Delli Bovi, Alessandro Raganato and Roberto Navigli}

\address{Department of Computer Science, Sapienza University of Rome \\
         Via Regina Elena, 295 - 00161 Roma, Italy\\
         {\tt \{collados,dellibovi,raganato,navigli\}@di.uniroma1.it}\\}

\abstract{
    Linking concepts and named entities to knowledge bases has become a crucial Natural Language Understanding task. In this respect, recent works have shown the key advantage of exploiting textual definitions in various Natural Language Processing applications. However, to date there are no reliable large-scale corpora of sense-annotated textual definitions available to the research community. In this paper we present a large-scale high-quality corpus of disambiguated glosses in multiple languages, comprising sense annotations of both concepts and named entities from a unified sense inventory. Our approach for the construction and disambiguation of the corpus builds upon the structure of a large multilingual semantic network and a state-of-the-art disambiguation system; first, we gather complementary information of equivalent definitions across different languages to provide context for disambiguation, and then we combine it with a semantic similarity-based refinement. As a result we obtain a multilingual corpus of textual definitions featuring over 38 million definitions in 263 languages, and we make it freely available at \url{http://lcl.uniroma1.it/disambiguated-glosses}. Experiments on Open Information Extraction and Sense Clustering show how two state-of-the-art approaches improve their performance by integrating our disambiguated corpus into their pipeline. \\
\newline 
}

\begin{document}

\maketitleabstract

\section{Introduction}
\setlength{\parindent}{1em}
In addition to lexicography, where their use is of paramount importance, textual definitions drawn from dictionaries or encyclopedias have been widely used in various Natural Language Processing tasks and applications. Some of the areas where the use of definitional knowledge has proved to be key in achieving state-of-the-art results are Word Sense Disambiguation \cite{Lesk:86,BanerjeePedersen:2002,NavigliVelardi:05,agirrepersonalizing:2009,fernandez2012unsupervised,chenunified:2014,camachocolladosetal:2015b}, Taxonomy and Ontology Learning \cite{Velardietal:2013cl,Flatietal:2014,espinosa2016extasem}, Information Extraction \cite{Richardsonetal:1998,Dellibovietal:15tacl}, Plagiarism Detection \cite{franco2016systematic}, and Question Answering \cite{hill2015learning}. \\\indent
In fact, textual definitions (or glosses) are today widely to be found in resources of various kinds, from lexicons and dictionaries, such as WordNet \cite{Milleretal:90} or Wiktionary, to encyclopedias and knowledge bases, such as Wikidata and OmegaWiki. These include Wikipedia itself: indeed, the first sentence of a Wikipedia article is generally regarded as the definition of its subject\footnote{According to the Wikipedia guidelines an article should begin with a short declarative sentence defining what (or who) the subject is and why it is notable.}. In any case, an accurate semantic analysis of a definition corpus is made difficult by the short and concise nature of definitional text. Furthermore, the majority of approaches making use of definitions are restricted to corpora where each concept or entity is associated with a single definition, while definitions coming from different resources are often complementary and might give different perspectives on the definiendum. Moreover, equivalent definitions of the same concept or entity may vary substantially according to the language, and be more precise or self-explanatory in some languages than others. This has the potential to be especially valuable in the context of disambiguation \cite{navigli:09}, where highly ambiguous terms in one language may become less ambiguous (or even unambiguous) in other languages.\\\indent
In this paper we bring together definitions coming from both different resources and different languages, and disambiguate them by exploiting their cross-lingual and cross-resource complementarities. Our goal is to obtain a large-scale high-quality corpus of sense-annotated textual definitions. In order to do this we leverage BabelNet\footnote{\url{http://babelnet.org}} \cite{NavigliPonzetto:12aij}, a multilingual lexicalized semantic network obtained from the automatic integration of lexicographic and encyclopedic resources. Due to its wide coverage of both lexicographic and encyclopedic terms, BabelNet gives us a very large sense inventory for disambiguation, as well as a vast and comprehensive target corpus of textual definitions. In fact, as it is a merger of various different resources, BabelNet provides a large heterogeneous set of over 35 million definitions for over 250 languages from WordNet, Wikipedia, Wiktionary, Wikidata and OmegaWiki. To the best of our knowledge, this set constitutes the largest available corpus of definitional text.\\\indent
We evaluate our sense-annotated corpus intrinsically, obtaining a disambiguation precision of over 90\% on a random sample of definitions in three different languages, and extrinsically on Open Information Extraction and Sense Clustering tasks. Our experiments show the potential of exploiting our disambiguated glosses within the pipelines of two state-of-the-art systems, improving on their original performance.

\medskip
\section{Related Work}

\paragraph{}Among all resources using textual definitions, WordNet has definitely been the most popular and the most exploited to date. In fact, WordNet glosses have still been used successfully in recent work \cite{kahn2013,chen2015improving}. \\ \indent 
A first attempt to disambiguate WordNet glosses automatically was proposed as part of the eXtended WordNet project\footnote{\url{http://www.hlt.utdallas.edu/~xwn/}}~\cite{novischi2002accurate}. However, this attempt's estimated coverage did not reach 6\% of the total amount of sense-annotated instances. Moldovan and Novischi~\shortcite{moldovan2004word} proposed an alternative disambiguation approach, specifically targeted at the WordNet sense inventory and based on a supervised model trained on the SemCor sense-annotated corpus~\cite{miller93}. In general, the drawback of supervised models arises from the so-called \textit{knowledge-acquisition bottleneck}, a problem that becomes particularly vexed when such models are applied to larger inventories, due to the vast amount of annotated data they normally require. Another disambiguation task focused on WordNet glosses was presented as part of the SensEval-3 workshop \cite{litkowski2004senseval}. However, the best reported system obtained precision and recall figures below 70\%, which arguably is not enough to provide high-quality sense-annotated data for current state-of-the-art NLP systems.\\ \indent 
In addition to annotation reliability, another issue that arises when producing a corpus of textual definitions is wide coverage. In fact, reliable corpora of sense-annotated definitions produced to date, such as the Princeton WordNet Gloss Corpus\footnote{\url{http://wordnet.princeton.edu/glosstag.shtml}}, have usually been obtained by relying on human annotators. The Princeton corpus of WordNet disambiguated glosses has already been shown to be successful as part of the pipeline in semantic similarity \cite{Pilehvaretal:2013}, domain labeling \cite{gonzalez2012graph} and Word Sense Disambiguation \cite{agirrepersonalizing:2009,camachocolladosetal:2015b} systems. However, as new encyclopedic knowledge about the world is constantly being harvested, keeping up using only human annotation is becoming an increasingly expensive endeavor. With a view to tackling this problem, a great deal of research has recently focused on the automatic extraction of definitions from unstructured text \cite{NavigliVelardi:10,debenedictisetal:2013,espinosaetal:2014,dalvietal:2015}. On the other hand, the prominent role of collaborative resources~\cite{Hovyetal:13} has created a convenient development ground for NLP systems based on encyclopedic definitional knowledge. Nevertheless, extending the manual annotation of definitions to much larger and up-to-date knowledge repositories like BabelNet is not feasible. First of all, the number of items to disambiguate is massive; moreover, as the number of concepts and named entities increases, annotators would have to deal with the added difficulty of selecting context-appropriate synsets from an extremely large sense inventory. In fact, WordNet 3.0 comprises 117 659 synsets and a definition for each synset, while BabelNet 3.0 covers 13 801 844 synsets with a total of 40 328 194 definitions.\\ \indent
Instead, in this paper we propose an automatic disambiguation approach which leverages multilinguality and cross-resource information along with a state-of-the-art multilingual Word Sense Disambiguation/Entity Linking system \cite{Moroetal:14tacl} and a vector-based semantic representation of concepts and entities \cite{camachocolladosetal:2015}. By exploiting these features, we are able to produce a large-scale high-quality corpus of glosses, automatically disambiguated with BabelNet synsets\footnote{Note that BabelNet covers WordNet and Wikipedia among other resources, which makes our sense annotations expandable to any of these resources.}.

\section{Methodology}
\label{methodology}

The gist of our approach lies in the combination of different languages and resources for high-quality disambiguation. In fact, since many definitions are short and concise, the lack of meaningful context would negatively affect the performance of a Word Sense Disambiguation/Entity Linking system targeted at individual definitions.

To improve the data quality before the disambiguation step, we tokenize and Part-of-Speech (PoS) tag the definitions for a subset of languages:
\paragraph{Tokenization.} We use the tokenization system available from the polyglot project\footnote{\url{http://polyglot.readthedocs.org/en/latest/Tokenization.html}} for 165 languages.
\paragraph{Part-of-Speech tagging.} We train the Stanford tagger \cite{toutanova2003feature}, for 30 languages using the available data from the Universal Dependencies project\footnote{\url{https://universaldependencies.github.io/docs/}}\cite{nivre2015towards}.

Our disambiguation strategy is based on two steps:  (1) all definitions are gathered together, grouped by definiendum and disambiguated using a multilingual disambiguation system (Section \ref{babelfy}); (2) the disambiguation output is then refined using semantic similarity (Section \ref{nasari}).

\subsection{Context-rich Disambiguation}  
\label{babelfy}

\paragraph{}As an example, consider the following definition of \textit{castling} in chess as provided by WordNet:
\begin{equation}
\label{example}
    \textit{Interchanging the positions of the king and a rook.} 
\end{equation}
The context in (\ref{example}) is limited and it might not be obvious for an automatic disambiguation system that the concept being defined relates to \textit{chess}: an alternative definition of \textit{castling} where the game of \textit{chess} is explicitly mentioned would definitely help the disambiguation process.
Following this idea, given a BabelNet synset, we carry out a \textit{context enrichment} procedure by collecting all the definitions of this synset in every available language and resource, and gathering them together into a single multilingual text. 

\begin{table*}[t]
\begin{center}
\resizebox{15.0cm}{!}{
    \centering
    \begin{tabular}{| c | r || r | r | r | r | r |}
        \cline{2-7}
          \multicolumn{1}{c|}{} & \multicolumn{1}{c||}{\textbf{All languages}} & \multicolumn{1}{c|}{\textbf{English}} & \multicolumn{1}{c|}{\textbf{Spanish}} & \multicolumn{1}{c|}{\textbf{French}} & \multicolumn{1}{c|}{\textbf{Italian}} &  \multicolumn{1}{c|}{\textbf{Persian}} \\
        \hline
        \small \textbf{Wikipedia} & \small 29 792 245 & \small 4 854 598 & \small 1 152 271 & \small 1 590 767 & \small 1 113 357 & \small 414 950 \\
        \hline
        \small \textbf{Wikidata} & \small 8 484 267 & \small 703 369 & \small 232 091 & \small 1 392 718 & \small 987 912 & \small 352 697 \\
        \hline
        \small \textbf{Wiktionary} & \small 281 756 & \small 281 756 & \small - & \small - & \small - & \small - \\
        \hline
        \small \textbf{OmegaWiki} & \small 115 828 & \small 29 863 & \small 22 446 & \small 12 777 & \small 14 763 & \small 11 \\
        \hline
        \small \textbf{WordNet} & \small 146 018 & \small 117 226 & \small - & \small - & \small - & \small - \\
        \hline
        \hline
        \small \textbf{Total} & \small \textbf{38 820 114} & \small \textbf{5 986 812} & \small \textbf{1 406 808} & \small \textbf{2 996 262} & \small \textbf{2 116 032} & \small \textbf{767 658} \\
        \hline
    \end{tabular}
     }
     \end{center}
    \caption{Number of disambiguated glosses by language (\textit{columns}) and by resource (\textit{rows}).}
    \bigskip
    \label{tab:glosses}

\end{table*}

We use a state-of-the-art graph-based approach to Entity Linking and Word Sense Disambiguation, Babelfy\footnote{\url{http://babelfy.org}} \cite{Moroetal:14tacl}, to disambiguate definitions after preprocessing and context-enrichment.
Our methodology relies on the fact that, as shown in Section \ref{babelfy}, disambiguation systems like Babelfy work better with richer context. When provided with the definition of Example (\ref{example}) in isolation, Babelfy incorrectly disambiguates \textit{rook} as "rookie, inexperienced youth". However, by using additional definitions from other resources and languages, Babelfy exploits the added context and disambiguates \textit{rook} with its correct \textit{chess} sense. 
This approach is particularly advantageous for languages with low resources, where standard disambiguation techniques have not yet proved to be reliable, due to the shortage of annotated data.

\subsection{Disambiguation Refinement}  
\label{nasari}

Babelfy outputs a set $D$ of \textit{disambiguated instances}, i.e. mappings from text fragments to items in the BabelNet sense inventory, each associated with a confidence score (\textit{Babelfy score} henceforth). When Babelfy score goes below 0.7, a back-off strategy based on the \textit{most common sense} is used by default for that instance. Our aim is to correct or discard these low-confidence instances using Semantic Similarity.\\\indent  
First, for each disambiguated instance $d \in D$ we compute a \textit{coherence score} $C_{d}$. The coherence score is provided by Babelfy as the number of semantic connections from $d$ to the rest of disambiguation instances in the semantic graph (normalized):
\begin{equation}
C_d=\frac{|\text{Disambiguated instances connected to $d$}|}{|\text{Disambiguated instances}| -1}
\end{equation}

We empirically set a coherence score threshold to 0.125 (i.e. one semantic connection out of eight disambiguated instances). Let $L$ be the set of disambiguated instances below both Babelfy and coherence score thresholds (low confidence). In order to refine the disambiguated instances in $L$, we use \textsc{NASARI}\footnote{We use the 2.1 release version of the \textit{NASARI-embed} vectors, downloaded from \url{http://lcl.uniroma1.it/nasari}} \cite{camachocolladosetal:2015,camachocolladosetal:2015b}. \textsc{NASARI} provides vector representations for over four million BabelNet synsets built by exploiting the complementary knowledge of Wikipedia and WordNet. These semantic representations have proved capable of obtaining state-of-the-art results in various lexical semantics tasks such as Semantic Similarity, Sense Clustering and Word Sense Disambiguation.
We consider those instances in $L$ for which a \textsc{NASARI} vector can be retrieved (virtually all noun instances), and compute an additional score (\textit{NASARI score}). First, we calculate the centroid $\mu$ of all the \textsc{NASARI} vectors for instances in $D \setminus L$. Then, for each disambiguated instance $l \in L$, we retrieve all the candidate senses of its surface form in BabelNet and calculate a NASARI score $N_s$ for each candidate sense. $N_s$ is calculated as the cosine similarity between the centroid $\mu$ and its corresponding \textsc{NASARI} vector $NASARI(s)$:
\begin{equation}
N_s=Sim(\mu,NASARI(s))
\end{equation}
The NASARI score allows us to both discard low-confidence disambiguated instances and correct the original disambiguation output by Babelfy in some cases. Then, each $l \in L$ is re-tagged with the sense obtaining the highest \textsc{NASARI} score:
\begin{equation}
\hat{s}=\argmax_{s \in S_l} {N_l}
\end{equation}
where $S_l$ is the set containing all the candidate senses for $l$. For what concerns the high-precision disambiguated glosses release (see Section \ref{release}) we set the NASARI threshold to 0.75. Considering example (\ref{example}) again, Babelfy does not provide a high-confidence disambiguation for the word \textit{king}, which is then incorrectly disambiguated using the most common sense strategy. However, the error is fixed during the refinement step: our system accurately selects the \textit{chess} sense of \textit{king} thanks to its high semantic connection with the disambiguated instances in $D\setminus L$. 

\begin{table*}[t]
\begin{center}
    \resizebox{16.0cm}{!}{
    \begin{tabular}{| l | c | r || r | r | r | r | r |}
        \cline{2-8}
         \multicolumn{1}{c|}{$\quad$} & \multicolumn{1}{c|}{} & \multicolumn{1}{c||}{\textbf{All languages}} & \multicolumn{1}{c|}{\textbf{English}} & \multicolumn{1}{c|}{\textbf{Spanish}} & \multicolumn{1}{c|}{\textbf{French}} & \multicolumn{1}{c|}{\textbf{Italian}} & \multicolumn{1}{c|}{\textbf{Persian}} \\
        \hline
         \multirow{3}{*}{\textbf{Before refinement}} 
         &\textbf{Babelfy} &  174 256 335 &  39 096 127 &  9 006 888 &  11 178 328 &  8 892 763 &  3 766 754\\
         \cline{2-8}
           \cline{2-8}
         &\textbf{MCS} &  75 288 373 &  19 724 340 &  5 164 557 &  7 064 210 &  4 525 610 &  1 524 267\\
         \cline{2-8}
         \cline{2-8}
         &\textbf{Total} &  \textbf{249 544 708} &  \textbf{58 820 467} &  \textbf{14 171 445} &  \textbf{18 242 538} &  \textbf{13 418 373} &  \textbf{5 291 021} \\
        \hline
        \hline
         \multirow{3}{*}{\textbf{After refinement}} 
         & \textbf{Babelfy}&  144 637 032 &  33 260 600 &  7 029 173 &  8 735 298 & 7 106 414 & 3 085 804 \\
         \cline{2-8}
         & \textbf{NASARI}&  18 392 099 &  4 680 745 &  1 353 494 &  1 865 920 &  1 301 370 & 330 917 \\
        \cline{2-8}
        \cline{2-8}
        & \textbf{Total} &  \textbf{163 029 131} &  \textbf{37 941 345} &  \textbf{8 382 667} &  \textbf{10 601 218} &  \textbf{8 407 784} &  \textbf{3 416 721} \\
        
        \hline
    \end{tabular}
}

\end{center}
    \caption{Number of annotations by language (\textit{columns}) and by type (\textit{rows}) before and after refinement.}
    \label{tab:annotations}
\end{table*}

\begin{table*}[t]
\begin{center}
    \resizebox{15.5cm}{!}{
    \begin{tabular}{| l | c | r || r | r | r | r | }
        \cline{2-7}
         \multicolumn{1}{c|}{$\quad$} & \multicolumn{1}{c|}{} & \multicolumn{1}{c||}{\textbf{All content words}} & \multicolumn{1}{c|}{\textbf{Nouns}} & \multicolumn{1}{c|}{\textbf{Verbs}} & \multicolumn{1}{c|}{\textbf{Adjectives}} & \multicolumn{1}{c|}{\textbf{Adverbs}}  \\
        \hline
         \multirow{3}{*}{\textbf{Before refinement}} 
         &\textbf{Babelfy} &   174 256 335 &  158 310 414 &  4 368 488 &  10 646 921 &  930 512 \\
         \cline{2-7}
           \cline{2-7}
         &\textbf{MCS} &  75 288 373 &  56 231 910 &  8 344 930 &  9 256 497 &  1 455 036 \\
         \cline{2-7}
         \cline{2-7}
         &\textbf{Total} &  \textbf{249 544 708} &  \textbf{214 542 324} &  \textbf{12 713 418} &  \textbf{19 903 418} &  \textbf{2 385 548} \\
        \hline
        \hline
         \multirow{3}{*}{\textbf{After refinement}} 
         & \textbf{Babelfy}&   144 637 032 &  140 111 921 &  1 326 947 &  3 064 416 &  133 748  \\
         \cline{2-7}
         & \textbf{NASARI}&  18 392 099 &  18 392 099 &  - &  - &  -  \\
        \cline{2-7}
        \cline{2-7}
        & \textbf{Total} &  \textbf{163 029 131} &  \textbf{158 504 020} &  \textbf{1 326 947} &  \textbf{3 064 416} &  \textbf{133 748}  \\
        
        \hline
    \end{tabular}
}

\end{center}
    \caption{Number of annotations by Part-of-Speech (PoS) tag (\textit{columns}) and by type (\textit{rows}) before and after refinement.}
    \label{tab:annotationsPOS}
\end{table*}

\section{Statistics}
\label{statistics}

The output of our disambiguation procedure is a corpus of 38 820 114 glosses extracted from BabelNet (corresponding to 8 665 300 BabelNet synsets), covering 263 languages and 5 different resources (Wiktionary, WordNet\footnote{Including Open Multilingual WordNet.}, Wikidata, Wikipedia\footnote{Definitions from Wikipedia include both first sentences of Wikipedia articles and definitions coming from Wikipedia's disambiguation pages.} and OmegaWiki) and including 249 544 708 annotations from the BabelNet sense inventory (6.4 annotations per definition on average). Table \ref{tab:glosses} reports some general statistics of the complete corpus of disambiguated textual definitions and for five sample languages: English, Spanish, French, Italian and Persian. 

The number of disambiguated instances, before and after the refinement step, are displayed in Tables \ref{tab:annotations} and \ref{tab:annotationsPOS}, organized, respectively, by language and Part-of-Speech (PoS). \textit{Babelfy} and \textit{NASARI} refer to the instances disambiguated by the two respective approaches and \textit{MCS} to the instances which were disambiguated using the Most Common Sense (MCS) heuristic. After refinement, 24.7\% of the low-confidence noun annotations are fixed using semantic similarity (see Section \ref{nasari}). Assuming the coverage of our first disambiguation step (see Section \ref{babelfy}) to be 100\%\footnote{There is no straightforward way to estimate the coverage of a disambiguation system automatically. In our first step using Babelfy, we provide disambiguated instances for all content words (including multi-word expressions) from BabelNet and also for overlapping mentions. Therefore the output of our first step, even if not perfectly accurate, may be considered to have full coverage in comparison with our refinement step. }, the coverage of our system after the refinement step is estimated to be 65.3\%. As shown in Table \ref{tab:annotationsPOS}, discarded annotations mostly include verbs, adjectives and adverbs, often harder to disambiguate as they are not directly related to the definiendum. In fact, the coverage of noun instances after refinement is estimated to be 73.9\%.


\section{Evaluation}
\label{evaluation}

\subsection{Intrinsic evaluation}
\label{intrinsic_evaluation}

We first carry out an intrinsic evaluation of the resource, by manually assessing the quality of disambiguation on some randomly extracted samples of definitions. We rely on three human judges and evaluate samples of 100 items for three languages. 
We evaluated the disambiguation output before and after the refinement step, and compared against a baseline where each definition is disambiguated in isolation with Babelfy. Table \ref{tab:evaluation} reports the evaluation on the three sample languages: English, Spanish and Italian. Although the disambiguation of context-free definitions improves only slightly with respect to the disambiguation of definitions in isolation, this improvement is consistent across languages. Furthermore, our system significantly increases the precision after the refinement step. Refinement reduces the coverage by 35\% for English, and by 43\% for Spanish and Italian, but increases precision by almost 11\% for English, 20\% for Spanish and 13\% for Italian. 

\begin{table}[h]
\begin{center}
\resizebox{8.4cm}{!}{
\centering

\begin{tabular}{|c | c|c | c|c | c|c |}
\cline{2-7}
        \multicolumn{1}{c|}{}  &  \multicolumn{2}{c|}{\textbf{English}} & \multicolumn{2}{c|}{\textbf{Spanish}} & \multicolumn{2}{c|}{\textbf{Italian}}   \\
        \cline{2-7}
\multicolumn{1}{c|}{} &                                {\bf Prec.}   & {\bf Cov.}  & {\bf Prec.}   & {\bf Cov.} &                                {\bf Prec.}   & {\bf Cov.}  
\\
         
    \hline    \textbf{Definitions in isolation}  &   84.2 & \bf 100 & 74.6 & \bf 100 &  77.6 & \bf 100 \\
        \hline
       \textbf{Context-rich defs. pre-refin.}  &  84.3 &  \bf 100 &  74.7 &  \bf 100 &  78.0 & \bf 100 \\
        \hline
       
       \textbf{Context-rich defs. post-refin.} &   \bf 95.1 & 64.8  & \bf 95.0  &   57.3 &  \bf 91.1 & 56.8 \\
       \hline
\end{tabular} 
}
\end{center}
\caption{Disambiguation precision (Prec.) and coverage (Cov.) percentage (\%) of the three different disambiguation strategies on the 300 sample definitions.}
\label{tab:evaluation}
\end{table}

\subsection{Extrinsic evaluation}
\label{extrinsic_evaluation}

The sense-annotated corpus of definitions is also evaluated extrinsically with two experiments. The first experiment (Section \ref{informationextraction}) evaluates our corpus before the high-precision refinement, and is focused on \textsc{DefIE} \cite{Dellibovietal:15tacl}, an Open Information Extraction (OIE) system that works on textual definitions. In its original implementation \textsc{DefIE} uses Babelfy to disambiguate definitions one-by-one before extracting relation instances. We modified that implementation and used the glosses disambiguated with our approach as input for the system, and we compared the extracted information with the information obtained by the original implementation.
The second experiment (Section \ref{senseclustering}), instead, evaluates our refined high-precision corpus, and focuses on the semantic representations of \textsc{NASARI} (Section \ref{nasari}). These representations were constructed based on the BabelNet semantic network. We reimplemented \textsc{NASARI} using the same network enriched with the high-precision disambiguated glosses and compared these with the original glosses in the sense clustering task.

\subsubsection{Open Information Extraction}
\label{informationextraction}

In this experiment we investigated the impact of our disambiguation approach on the definitional corpus used as input for the pipeline of \textsc{DefIE}. The original OIE pipeline of the system takes as input an unstructured corpus of textual definitions, which are then preprocessed one-by-one to extract syntactic dependencies and disambiguate word senses and entity mentions. After this preprocessing stage, the algorithm constructs a syntactic-semantic graph representation for each definition, from which subject-verb-object triples (relation instances) are eventually extracted. As highlighted in Section \ref{babelfy}, poor context of particularly short definitions may introduce disambiguation errors in the preprocessing stage, which then tend to propagate and reflect on both relations and relation instances. To assess the quality of our disambiguation methodology as compared to a standard approach, we modified the implementation of \textsc{DefIE} to consider our disambiguated instances instead of executing the original disambiguation step, and then we evaluated the results obtained at the end of the pipeline in terms of quality of relation and relation instances.
\paragraph{Experimental setup.}
We first selected a random sample of 150 textual definitions from our disambiguated corpus (Section \ref{statistics}). We generated a baseline for the experiment by discarding all disambiguated instances from the sample, and treating the sample itself as an unstructured text of textual definitions which we used as input for \textsc{DefIE}, letting the original pipeline of the system carry out the disambiguation step. Then we carried out the same procedure using, instead, the modified implementation for which our disambiguated instances are taken into account. In both cases, we ran the extraction algorithm of \textsc{DefIE} and evaluated the output in terms of both relations and relation instances. Following Delli Bovi et al.~\shortcite{Dellibovietal:15tacl}, we relied on two human judges and performed the same evaluation procedure described therein over the set of distinct relations extracted from the sample, as well as the set of extracted relation instances.
\begin{table}[t]
\resizebox{7.8cm}{!}{
\centering
\begin{tabular}{| l | c | c | c |}
\cline{2-4}
\multicolumn{1}{c|}{} & {\bf \# Glosses} & {\bf \# Triples} & {\bf \# Relations}\\
    \hline
    \textbf{\textsc{DefIE} + glosses$\qquad$} & \textbf{150} & \textbf{340} & \textbf{184}\\ 
    \hline
    \textbf{\textsc{DefIE}}	& 146 & 318 & 171 \\ 
    \hline
\end{tabular}
}
\caption{\small \label{tab:defie} Extractions of \textsc{DefIE} on the evaluation sample.}
\end{table}
\begin{table}[t]
\resizebox{7.8cm}{!}{
\centering
\begin{tabular}{| l | c | c |}
\cline{2-3}
\multicolumn{1}{c|}{} & {\bf Relation} & {\bf Relation Instances} \\
    \hline
    \textbf{\textsc{DefIE} + glosses$\qquad$} & \textbf{0.872} & \textbf{0.780}\\ 
    \hline
    \textbf{\textsc{DefIE}}	& 0.865 & 0.770\\ 
    \hline
\end{tabular}
}
\caption{\small \label{tab:defie2} Precision of \textsc{DefIE} on the evaluation sample.}
\end{table}
\paragraph{Results.}
Results reported in Tables \ref{tab:defie} and \ref{tab:defie2} show a slight but consistent improvement resulting from our disambiguated glosses over both the number of extracted relations and triples and over the number of glosses with at least one extraction (Table \ref{tab:defie}), as well as over the estimated precision of such extractions (Table \ref{tab:defie2}). Context-rich disambiguation of glosses across resources and languages enabled the extraction of 6.5\% additional instances from the sample (2.26 extractions on the average from each definition) and, at the same time, increased the estimated precision of relation and relation instances over the sample by $\sim$1\%.

\subsubsection{Sense Clustering}
\label{senseclustering}

This experiment focuses on the sense clustering task. Knowledge resources such as Wikipedia or WordNet suffer from the high granularity of their sense inventories. A meaningful cluster of senses within these sense inventories would help boost the performance in different applications \cite{Hovyetal:13}. In this section we will explain how to deal with this issue in Wikipedia. 

We integrate the high-precision version of the network as enrichment of the BabelNet semantic network in order to improve the results of the state-of-the-art system based on NASARI lexical vectors (more details of NASARI in Section \ref{nasari}). NASARI uses Wikipedia ingoing links and the BabelNet taxonomy in the process of obtaining contextual information for a given concept. We simply enrich  the BabelNet taxonomy with the high-precision disambiguated glosses (see Section \ref{nasari}) of the target language. The high-precision disambiguated glosses are synsets that are highly semantically connected with the definiendum, which makes them particularly suitable for enriching a semantic network. The rest of the default NASARI lexical pipeline for obtaining semantic representations (lexical specificity applied to the contextual information) remains unchanged.  By integrating the high-precision disambiguated glosses into the NASARI pipeline, we obtain a new set of vector representations for BabelNet synsets, increasing its initial coverage (4.4M synsets covered by the default NASARI compared to 4.6M synsets covered by NASARI enriched with our disambiguated glosses). 

\paragraph{Experimental setup.}
We used the two sense clustering datasets created by \newcite{Dandalaetal:2013}. The task in these datasets consists of, given a pair of Wikipedia articles, to decide whether they should be merged into a single cluster or not. The first dataset (\textit{500-pair} henceforth) contains 500 pairs of Wikipedia articles, while the second dataset (\textit{SemEval}) consists of 925 pairs coming from a set of highly ambiguous words taken from disambiguation tasks of SemEval workshops. We follow the original setting of \cite{camachocolladosetal:2015} and only cluster a pair of Wikipedia articles if their similarity, calculated by using the square-rooted Weighted Overlap comparison measure \cite{Pilehvaretal:2013}, surpasses 0.5 (i.e. the middle point in the Weighted Overlap similarity scale).

\paragraph{Results.}
Table \ref{tab:clustering} shows the results of different systems in the sense clustering task. As a naive baseline we include a system which clusters all pairs. For comparison we also include the Support Vector Machine classifier of \newcite{Dandalaetal:2013} exploiting information of Wikipedia in four different languages (\textit{Dandala-multilingual}). Finally, we report the results of the default NASARI English lexical vectors (\textit{NASARI}\footnote{Downloaded from \url{http://lcl.uniroma1.it/nasari/}}) and the NASARI-based vectors obtained from the BabelNet semantic network enriched with our high-precision disambiguated glosses (\textit{NASARI+glosses}). As we can see from Table \ref{tab:clustering}, the enrichment produced by our glosses proved to be highly beneficial, significantly improving on the original results obtained by NASARI. Moreover, NASARI+glosses obtains the best performance overall, outperforming Dandala-multilingual in terms of accuracy in both datasets. 

\begin{table}[h]
\begin{center}
\resizebox{7.5cm}{!}{
\centering

\begin{tabular}{|c | c|c | c|c  |}
\cline{2-5}
        \multicolumn{1}{c|}{}  &  \multicolumn{2}{c|}{\textbf{500-pair}} & \multicolumn{2}{c|}{\textbf{SemEval}} \\
        \cline{2-5}
\multicolumn{1}{c|}{} &     {\bf Acc.}   & {\bf F1}  & {\bf Acc.}    &  {\bf F1}	
\\
         
    \hline \textbf{ \textsc{NASARI}+Glosses  }	&	\bf 86.0	& \bf 74.8	&	\bf 88.1	& \bf 64.7	\\ 
        \hline
     \textbf{  \textsc{NASARI} }	&		81.6				& 65.4	&	85.7	& 57.4	\\
        \hline
    \textbf{   Dandala-multilingual}               			&	 84.4	& - &	85.5	& - \\
       \hline
\textbf{Baseline  }       	&	28.6		& 44.5	&	17.5	& 29.8 \\
       \hline
\end{tabular} 
}
\end{center}
\caption{\label{tab:clustering} Accuracy (Acc.) and F-Measure (F1) percentages of different systems on the Wikipedia sense clustering datasets.}
\end{table}

\begin{figure*}[t]
    \resizebox{0.6\textwidth}{!}{\begin{minipage}{\textwidth}
        \includegraphics[scale=0.65]{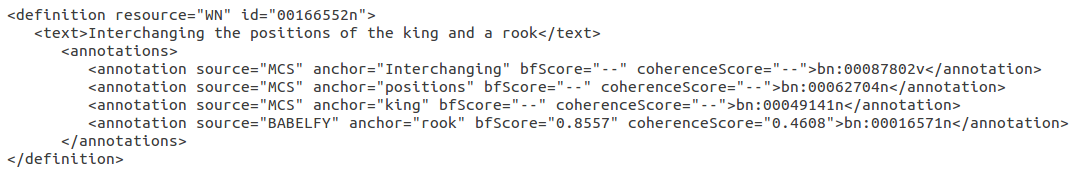}
    \end{minipage} }\\
        \caption{Sample XML output for the definition of \textit{castling} in WordNet from the complete disambiguated corpus.}
        \label{fig:castlingBF}
\end{figure*}

\begin{figure*}[t]
    \resizebox{0.6\textwidth}{!}{\begin{minipage}{\textwidth}
        \includegraphics[scale=0.65]{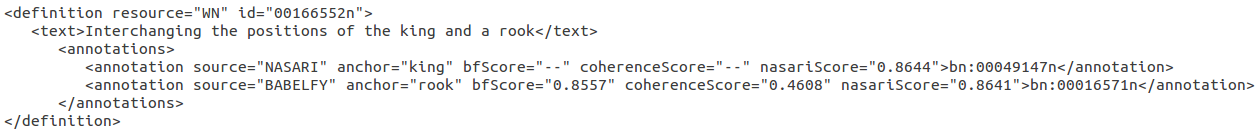}
    \end{minipage} }\\
        \caption{Sample XML output for the definition of \textit{castling} in WordNet from the high-precision disambiguated corpus.}
        \label{fig:castlingBFNS}
\end{figure*}

\section{Release}
\label{release}

The corpus of disambiguated glosses is freely available at \url{http://lcl.uniroma1.it/disambiguated-glosses}.
We released both the \textit{complete} (Section \ref{babelfy}) and the \textit{high-precision} (Section \ref{nasari}) versions of our corpus. 
The format for each of the two versions is almost identical: the corpus is first divided by resource (WordNet, Wikipedia, Wiktionary, Wikidata and OmegaWiki) and each resource is then divided by language.

The disambiguated glosses for each language and resource are stored in standard XML files. Figures \ref{fig:castlingBF} and \ref{fig:castlingBFNS} show a sample definition as displayed in the XML files of, respectively, the high-precision and complete version of our disambiguated corpus.
Each file contains a list of \texttt{definition} tags, with their respective \texttt{id}\footnote{Identifiers depend on the resource, e.g. offsets in WordNet and page titles in Wikipedia.} as attribute. Then, each definition tag is composed by the original definition as plain text and annotations.
The \texttt{annotation} tag refers to the sense-annotations provided as a result of our disambiguation process.
Each annotation includes its disambiguated BabelNet id and has four (or five) attributes (see Section \ref{methodology} for more details about the attributes):
\begin{itemize}
    \item \texttt{source}: this indicates whether the disambiguation has been performed by Babelfy, the Most Common Sense ("MCS") heuristic (only in the complete version of the corpus) or NASARI (only in the high-precision version of the corpus).
    \item \texttt{anchor}: this corresponds to the exact surface form match found within the definition.
    \item \texttt{bfScore}: this corresponds to the Babelfy score.
    \item \texttt{coherenceScore}: this corresponds to the coherence score.
    \item \texttt{nasariScore}: this corresponds to the NASARI score (only for the high-precision annotations).
\end{itemize}



\section{Conclusion}
\label{conclusion}

In this paper we presented a large-scale multilingual corpus of disambiguated glosses. Disambiguation was performed by exploiting cross-resource and cross-language complementarities of textual definitions. By leveraging the structure of a wide-coverage semantic network and sense inventory like BabelNet, we obtained a fully disambiguated corpus of textual definitions coming from multiple sources and multiple languages which, to the best of our knowledge, constitutes the largest available corpus of its kind. Additionally, we refined our sense annotations by integrating a module based on semantic similarity into our disambiguation pipeline, in order to identify a subset of high-precision disambiguated instances across the definitions. This refined version of the corpus has a great potential in high-precision low-coverage applications, where having a disambiguation error as low as possible is the first requirement. Since the disambiguated instances in this version of the corpus are directly connected to the definiendum, this high-precision disambiguated corpus may also be used to enrich a semantic network, or even used as a semantic network on its own.
We evaluated our corpus intrinsically on three different languages, showing that our system outperforms previous approaches and a standard state-of-the-art disambiguation system in terms of coverage, precision and recall. We also carried out an extrinsic evaluation that shows some applications of our resource: we integrated the complete and high-precision versions of our corpus into the pipeline of both an Open Information Extraction system and a Sense Clustering system, improving on their original results and obtaining state-of-the-art figures in both tasks.


\section*{Acknowledgments}

\vspace{1ex}
\noindent
\begin{minipage}{0.1\linewidth}
  \raisebox{-0.3\height}{\includegraphics[trim =32mm 55mm 30mm 5mm, clip, scale=0.23]{figs/erc.ai}}
\end{minipage}
\hspace{0.01\linewidth}
\begin{minipage}{0.72\linewidth}
  The authors gratefully acknowledge the support of the ERC Starting
  Grant MultiJEDI No.\ 259234.
\end{minipage}
\hspace{0.01\linewidth}
\begin{minipage}{0.05\linewidth}
  \vspace{0.05cm}
\raisebox{-0.25\height}{\includegraphics[trim =0mm 5mm 5mm 2mm,clip,scale=0.085]{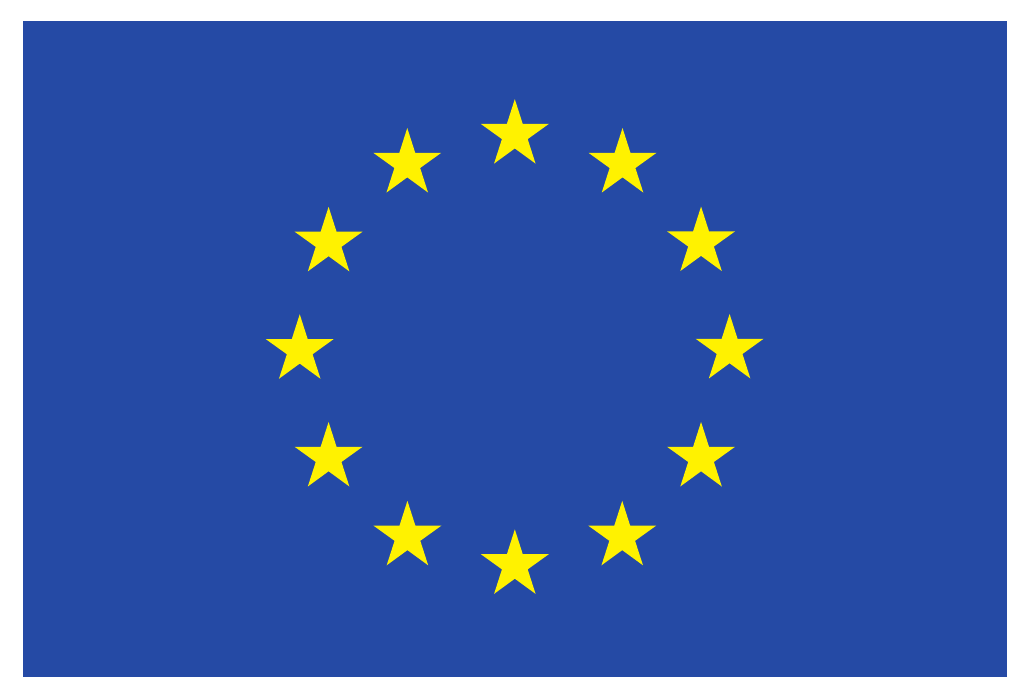}}
\end{minipage}
\vspace{2ex}

\section{References}
\label{main:ref}

\bibliographystyle{lrec2016}
\bibliography{xample}



\end{document}